\documentclass[a4paper]{article}[11pts]

\usepackage[english]{babel}
\usepackage[utf8x]{inputenc}
\usepackage[T1]{fontenc}

\normalsize

\usepackage[a4paper,top=1in,bottom=1in,left=1in,right=1in,marginparwidth=1in]{geometry}

\usepackage{setspace}
\usepackage{amsmath}
\usepackage{amssymb}
\usepackage{amsthm}
\usepackage{amsfonts, mathtools}
\usepackage{algorithm,algpseudocode}
\usepackage{bm}
\usepackage{bbm}
\usepackage{comment}
\usepackage{graphicx}
\usepackage{multirow, booktabs}
\usepackage{caption}
\usepackage{subcaption}
\usepackage[colorinlistoftodos]{todonotes}
\usepackage{url}

\newcommand{\prob}{\mathbb{P}}

\newtheorem{theorem}{Theorem}
\newtheorem{corollary}{Corollary}

\usepackage{natbib}

\title{Transformer Choice Net: A Transformer Neural Network for Choice Prediction}

\author{Hanzhao Wang \and Xiaocheng Li  \and  Kalyan Talluri}
\date{\small
Imperial College Business School, Imperial College London\\
$\{$h.wang19, xiaocheng.li, kalyan.talluri$\}$@imperial.ac.uk\\
}

\begin{document}
\maketitle

\onehalfspacing

\begin{abstract}

Discrete-choice models, such as Multinomial Logit, Probit, or Mixed-Logit, are widely used in Marketing, Economics, and Operations Research: given a set of alternatives, the customer is modeled as choosing one of the alternatives to maximize a (latent) utility function. However, extending such models to situations where the customer chooses more than one item (such as in e-commerce shopping) has proven problematic.  While one can construct reasonable models of the customer's behavior, estimating such models becomes very challenging  because of the combinatorial explosion in the number of possible subsets of items.  In this paper we develop  a transformer neural network architecture, the Transformer Choice Net, that is suitable for predicting multiple choices.  Transformer networks turn out to be especially suitable for this task as they take into account not only the features of the customer and the items but also the context, which in this case could be the assortment as well as the customer's past choices. On a range of benchmark datasets, our architecture shows  uniformly superior out-of-sample prediction performance compared to the  leading models in the literature, without requiring any custom modeling or tuning for each instance.
\end{abstract}

\section{Introduction}
Firms are interested in understanding the choice behavior of their customers as well as forecasting the sales of their items.  When customers choose at most one item per shopping instance, discrete-choice models estimate the probability of the choice, either at a segment level or individual customer level, based on a latent utility function of the features of the item, the customer, and the provided assortment.

However, there are many situations where customers choose multiple items on a single shopping instance, either from the same category or across categories. The firm may be aware of only the final choices made by the customer (as in physical retail) or the precise sequence of those choices  (such as in an e-commerce setting). 
\textit{Multi-choice models} are used for the former case, to estimate the probability of choosing a subset of items, amongst all possible subsets of the given assortment, considering potential interactions amongst the items and their features. 
\textit{Sequential choice models} consider the sequence of choices, taking into account not only the item and customer features but also what the customer has chosen till then to predict the subsequent choice(s).
 
Modeling and predicting the choice probabilities  for these situations is challenging: the complexity of the sequential and multi-choice models is considerably more than in the single-choice case because of combinatorial explosion in the number of possible customer journeys and final choices,  and consequently models for multiple choices are less widely adapted in practice.   

In this paper, we introduce the Transformer Choice Net, a neural network using the Transformer architecture \citep{vaswani2017attention}, as a data-driven solution that works under any of the three models: single, sequential, and multiple.

The key contributions of this paper are as follows:
\begin{itemize}
\item We develop the Transformer Choice Net, a neural network-based framework capable of encompassing all three choice paradigms---single, sequential, and multiple choices.  This unification simplifies the traditionally fragmented approach to choice modeling. 
\item We empirically validate the prediction performance of our model across diverse industry benchmark datasets. Our architecture shows uniformly superior out-of-sample prediction performance compared to the leading models in the discrete-choice literature, without requiring any custom modeling or tuning for each instance.  Moreover, our architecture is scalable, showing an ability to learn and predict multiple choices over a large assortment size, overcoming the limitations of many sequential and multi-choice models.  
\item We show theoretically the universal representation capacity of the Transformer Choice Net across the three choice paradigms.
\end{itemize}

%

\section{Literature Review}
\label{sec:review}
This section reviews the literature for single, sequential, and multiple choices, and also the literature on neural networks for predicting the choice probabilities.

\textbf{Single-choice models.} For general background on discrete-choice models, see \citet{mcfadden2000mixed} and the book by \citet{Ben-Akiva85}.

A plethora of models have been proposed for the single-choice case starting from the popular Multinomial Logit (MNL) choice model: to name a few, the Probit \citep{daganzo2014multinomial},  the ranking-based choice model \citep{block1959random, farias2013nonparametric}, the Negative-Exponential (also called Exponomial) model \citep{alptekinouglu2016exponomial}, and the Markov-chain choice model \citep{blanchet2016markov}.
 
From a machine learning perspective, the single-choice case can be viewed as a classification task where we assign probabilities for the various items (classes) to be chosen.  From this point of view, researchers have applied random forests \citep{chen2021estimating,chen2022decision}, embedding methods \citep{seshadri2019discovering}, and assortment context vectorization  \citep{yousefi2020choice,tomlinson2021learning,najafi2023assortment,bower2020preference} with the latter trying to overcome some known weaknesses of the MNL model.  \citet{van2022choice} provide a comprehensive review of the current literature on integrating machine learning into single-choice modeling.

\citet{bentz2000neural} give an early effort to use neural networks for single-choice modeling. \citet{wang2020deep,han2020neural,sifringer2020enhancing,wong2021reslogit,arkoudi2023combining} investigate different neural network architectures to capture the relationships between the items' features and utility in the MNL choice model. Several recent works try to capture more complicated assortment effects: the RUMNet \citep{aouad2022representing} mimics the mixed-MNL model instead of the MNL model, and the AssortNet \citep{wang2023neural} encodes the entire assortment into the choice probability via ResNet architecture. \citet{pfannschmidt2022learning,rosenfeld2020predicting} employ the Deep Set architecture \citep{zaheer2017deep} to allow permutation variations in input assortments (see also \cite{wagstaff2019limitations} and \S\ref{sec:TCN}).

\textbf{Sequential and multi-choice models.} Sequential choice is typically based on purchase history data, often from assortments with only a small number of items ($\le10$) \citep{manchanda1999shopping,gupta1988impact}. Despite the growth of e-commerce and the availability of click-stream data, it is somewhat surprising that there are relatively few models for sequential choice behavior:
The SHOPPER model \citep{10.1214/19-AOAS1265} stands out as a probabilistic representation of consumer choice, aiming to forecast sequential choices based on variational inference.  
Another noteworthy approach is due to \citet{jacobs2021understanding}, where they compress purchase history data into purchase motivations and then employ variational inference for model deductions.   \citet{gabel2022product} design a neural network that takes each customer's history  as input to predict the next choice, but like \cite{jacobs2021understanding} the architecture does not encode the customer and item features.

There is a slightly richer body of research on multiple choices, i.e., without knowing the sequence of choices~\citep{chen2023assortment,bai2023assortment,luan2023operations,jasin2023assortment,tulabandhula2023multi}.  \citet{benson2018discrete} expand the random utility maximization model from single to multi-choice scenarios. \citet{tulabandhula2023multi} take a different approach, quantifying pairwise item interactions within each potentially chosen subset through a linear sum.  \citet{aarts2023interpretable} bridge point processes with subset-choice models, encoding pairwise interactions using the    $\log\det$ function.  \citet{lin2022multi} augment the ranking-based choice model \citep{block1959random, farias2013nonparametric} to multiple choices, but they do not explicitly model item or customer features. A common limitation for all these models is scalability: as we have to predict the most likely amongst an exponential number of subsets, the size of the assortment they can handle is small.

\textbf{Transformer architecture.} First introduced by \citet{vaswani2017attention}, the Transformer architecture has emerged as a ground-breaking deep learning framework. Initially tested on machine translation within the realm of natural language processing (NLP), it has since expanded to a broad range of NLP tasks as expemplified by ChatGPT \citep{openai2023gpt4} for text generation. 

The Transformer architecture has also excelled in computer vision \citep{han2022survey}, audio processing \citep{dong2018speech}, and integrated multimodal tasks \citep{ramesh2021zero}. For a survey, we refer to the review by \citet{lin2022survey}.  Relevant to this paper, \citet{lee2019set} design the Set Transformer, which ensures any permutation of the input to yield identical outputs (see \S\ref{sec:TCN} for more discussions). To the best of our knowledge, there has been no application till now to choice modeling.

\section{Problem Setup}
\label{sec:setup}
\subsection{Preliminaries on Choice Models}
\subsubsection{Single-Choice Model}

Consider a ground set of $n$ potential items $\mathcal{N}=\{1,2,...,n\}$. For completeness, we can let $n$ denote the no-purchase option which (when selected) means the customer chooses none of the offered items. A seller chooses an \textit{assortment} $\mathcal{S}\subseteq \mathcal{N}$ to offer to the customer, and in the single-choice case the customer will choose at most one item from it. A \textit{single-choice model} prescribes the probability of choosing item $i$ conditional on the assortment $\mathcal{S}$ (where we can assume the no-purchase option is always in the assortment):
$$\prob(i|\mathcal{S}) \text{ for all } i\in\mathcal{N} \ \text{and} \ \mathcal{S}\subseteq \mathcal{N}.$$
In particular, $\prob(i|\mathcal{S})=0$ for $i\notin \mathcal{S},$ i.e., the customer cannot choose an item not offered in the assortment. In this way, a single-choice model
\begin{equation}
    \mathcal{M} = \{\prob(i|\mathcal{S}): \mathcal{S}\subseteq \mathcal{N}\}
    \label{choice_model}
\end{equation}
dictates $2^n-1$ probability distributions, each of which corresponds to one possible assortment (excluding when $\mathcal{S}$ is an empty set).

\subsubsection{Sequential Choice Model} The sequential choice model is an intermediate between the single-choice model and the multi-choice model, and it also serves as a useful model by itself. Consider a customer shopping at an e-commerce who may choose more than one item for purchase and the items are chosen sequentially one after another from an assortment. To capture this scenario, we define the set of currently not chosen items as \textit{candidates} (denoted by $\mathcal{C}$), while the set of initially offered items as \textit{assortment} (denoted as $\mathcal{S}$). Thus $\mathcal{C}\subseteq \mathcal{S}$ and the set $\mathcal{S}\setminus\mathcal{C}$ is the set of chosen items so far. The single-choice model then can be viewed as a special case where $\mathcal{C}=\mathcal{S}$. The sequential choice model describes the probability of choosing item $i$ from candidates $\mathcal{C}$ given the assortment $\mathcal{S}$
$$\prob(i|\mathcal{C},\mathcal{S}) \text{ for all } i\in\mathcal{N}, \mathcal{C}\subseteq  \mathcal{S}\subseteq \mathcal{N}.$$
Let  $\prob(i|\mathcal{C},\mathcal{S})=0$ for $i\notin \mathcal{C}$, i.e., the customer can only choose an item from the candidates set. In parallel to \eqref{choice_model}, the sequential choice model encapsulates the following distributions
\begin{equation}
    \mathcal{M} = \{\prob(i|\mathcal{C},\mathcal{S}): \mathcal{C}\subseteq \mathcal{S} \subseteq  \mathcal{N}\}.
    \label{seq_choice_model}
\end{equation}

\subsubsection{Multi-Choice Model} The multi-choice model considers the purchase of a basket of items, i.e., a subset of the given assortment. Hence, we are interested in the following distributions
\begin{equation}
    \mathcal{M} = \{\prob(\mathcal{B}|\mathcal{S}):  \mathcal{S}\subseteq \mathcal{N}\}
    \label{mul_choice_model}
\end{equation}
where $\mathcal{B}\subseteq \mathcal{N}$ denotes the basket of (chosen) items and $\mathcal{S}$ is the given assortment as before.  As earlier, we assume $\prob(\mathcal{B}|\mathcal{S})=0$ if $\mathcal{B}\nsubseteq \mathcal{S}$; in other words, the chosen basket can only contain items from the assortment.

\subsection{Learning Task}
\label{sec:learning}
The learning task, the focus of our paper, refers to estimating the choice model $\mathcal{M}$ from $m$ observed samples $\mathcal{D}$:
\begin{equation*}
\mathcal{D}=\begin{cases} \{(i_k, \mathcal{S}_k), k=1,...,m\} &\text{single choice},\\
\{(i_k, \{\mathcal{C}_k,\mathcal{S}_k\}), k=1,...,m\} &\text{sequential choice},\\
\{(\mathcal{B}_k,\mathcal{S}_k\}), k=1,...,m\} &\text{multiple choices},\\
\end{cases}
\end{equation*}
where each observation, depending on the model to be estimated, consists of the final choice(s) ($i_k$ or $\mathcal{B}_k$), accompanied by the assortment information ($\mathcal{S}_k$ or $\{\mathcal{C}_k,\mathcal{S}_k\}$).

\textbf{Item/product feature}. We assume that there is a feature vector associated with each item/product. Such a feature vector encodes characteristics of the product such as color, size, brand, price, etc. When no such feature is available, one can still obtain such a vector by a one-hot encoding of the items.

\textbf{Reduction to sequential choice model.} The goal of our paper is to develop one single transformer model that fits all three choice modeling scenarios. In this light, we discuss here how the single-choice model and the multi-choice model can be reduced to the sequential choice model in terms of both learning and inference so the transformer model can be based only on learning and predicting a sequential choice model.

The reduction of the single-choice model to the sequential choice model is straightforward. As we pointed out earlier, we simply set $\mathcal{C}=\mathcal{S}$, i.e., to predict the first item to be chosen given the initial assortment. For the estimation part, the sequential choice model can be estimated from the single-choice training samples $\{(i_k, \mathcal{S}_k), k=1,...,m\}$  by simply augmenting it into the form $\{(i_k, \{\mathcal{S}_k,\mathcal{S}_k\}), k=1,...,m\}$. For the inference part, one can infer the probabilities of the single-choice model by using the probability $\prob(i|\mathcal{S},\mathcal{S})$ from the sequential choice model.

The reduction from the multi-choice model to the sequential choice model is a bit more complicated. First, we note that the multi-choice model can also be mathematically represented by the sequential choice model. To see it, we can introduce a sequence $\sigma$ to denote the order of items being added to the basket. Then we can decompose the choice probability of a basket by sequential choice probabilities:
\begin{equation}
    \prob(\mathcal{B}|\mathcal{S})=\sum_{\sigma\in \text{Perm}_{\mathcal{B}}}\prod_{j=1}^{|\mathcal{B}|}\mathbb{P}(\sigma_j|\mathcal{C}^{\sigma,\mathcal{S}}_j,\mathcal{S}),
    \label{eqn:decompose_mul}
\end{equation}
where $\text{Perm}_{\mathcal{B}}$ contains all possible permutations over the set $\mathcal{B}$, $\sigma_j$ is the $j$-th item in the permutation $\sigma$, and $\mathcal{C}^{\sigma,\mathcal{S}}_j=\mathcal{S}\setminus\{\sigma_i:i<j \}$, i.e., the candidates when adding $j$-th item to the basket.

With this representation in mind, we first consider the learning of a sequential choice model from multi-choice training samples $\{(\mathcal{B}_k,\mathcal{S}_k\}), k=1,...,m\}$. Since the sequence in which the items are added to the basket $\mathcal{B}_k$ is unobserved, we cannot directly use the data to train a sequential choice model. One remedy involves an iteratively random selection of items from the basket, designating it as the sequential choice, and then constructing its associated candidates. For example, we are given a multi-choice sample $\{\mathcal{B},\mathcal{S}\}$ and we start the sample generation (for the sequential choice model) by setting  $\mathcal{C} = \mathcal{S}$. Say, the first randomly chosen item from $\mathcal{B}$ is $i^{(1)}$; then we obtain one sample of $\{i^{(1)}, (\mathcal{S}, \mathcal{S})\}$. Then we move on to generating the second sample, $\{i^{(2)}, (\mathcal{S}\backslash\{i^{(1)}\} , \mathcal{S})\}$ where $i^{(2)}$ is randomly picked from $\mathcal{B}\backslash\{i^{(1)}\} $. By continuing this process, we generate training samples for the sequential choice model from the original multi-choice sample $\{\mathcal{B},\mathcal{S}\}$. A similar sampling method is also applied in \citet{10.1214/19-AOAS1265}.

To predict the multiple choices of a given assortment $\mathcal{S}$ from a sequential choice model,  the set of predicted choices can be  sequentially generated by reversing the above sampling method. Specifically, we can start with $\mathcal{C}=\mathcal{S}$ and either sample a product $i^{(1)}$ following the distribution $\prob(\cdot|\mathcal{C},\mathcal{S})$ or select the item with the largest choice probability, and then remove $i^{(1)}$ from $\mathcal{C}$. By continuing this process, we can sample multiple items and add them to the predicted set of choices, until a stopping rule is met. This stopping rule can be enforced either when a pre-determined size of the basket is met or a virtual ``stopping item'' is sampled. 

\section{Transformer Choice Net}
\label{sec:TCN}
Now we introduce the Transformer Choice Net (TCNet) for the sequential choice model (which as we pointed out covers both single and multi-choices). Mathematically, the TCNet is denoted as $f^{\text{TCN}}$, a function that maps the candidate set $\mathcal{C}$ and the assortment $\mathcal{S}$ to the choice probability $f_i^{\text{TCN}}(\mathcal{C},\mathcal{S})$ for each item $i\in \mathcal{C}$.

\subsection{Attention Mechanism}

The attention mechanism has become a fundamental component of modern deep learning to solve various tasks of natural language processing and computer vision. In this subsection, we provide some preliminaries on this mechanism (see \citep{niu2021review} for more details) and show how it can be adapted for choice modeling.

An \textit{Attention} function  $\text{Att}(Q,K,V)$ takes three matrices as input: the matrix of queries $Q\in \mathbb{R}^{S\times d}$, the matrix of keys $K\in \mathbb{R}^{S'\times d}$, and the matrix of values $V\in \mathbb{R}^{S'\times d}$
$$\text{Att}(Q,K,V)=\phi(QK^\top)V,$$
where $d$ is the feature dimension.  The function $\phi$ is an activation function to $QK^\top$, with Softmax being the most prevalent choice. The matrix $QK^\top$ is termed the (unnormalized) score matrix, the $(i,j)$-element of which quantifies the influence of item $j$ on item $i$. Intuitively, the attention mechanism empowers a model to selectively focus on specific segments of its input, guided by the score function, to generate an output.

\textit{Self-attention} is an exemplification of the attention mechanism where all the three matrices above depend on the input feature matrix $X$:
\begin{equation}
f^{\text{Att}}(X;W_Q,W_K,W_V) = \text{Att}(XW_Q, XW_K, XW_V),
\label{eqn:self_att}
\end{equation}
where $ W_Q, W_K, W_V \in \mathbb{R}^{d\times d}$ are trainable parameters. For simplicity, we abbreviate the parameters and use $f^{\text{Att}}(X)$ to denote this self-attention function.

\subsection{Architecture of TCNet}

\begin{figure*}[ht!]
    \centering
    \includegraphics[scale=0.7]{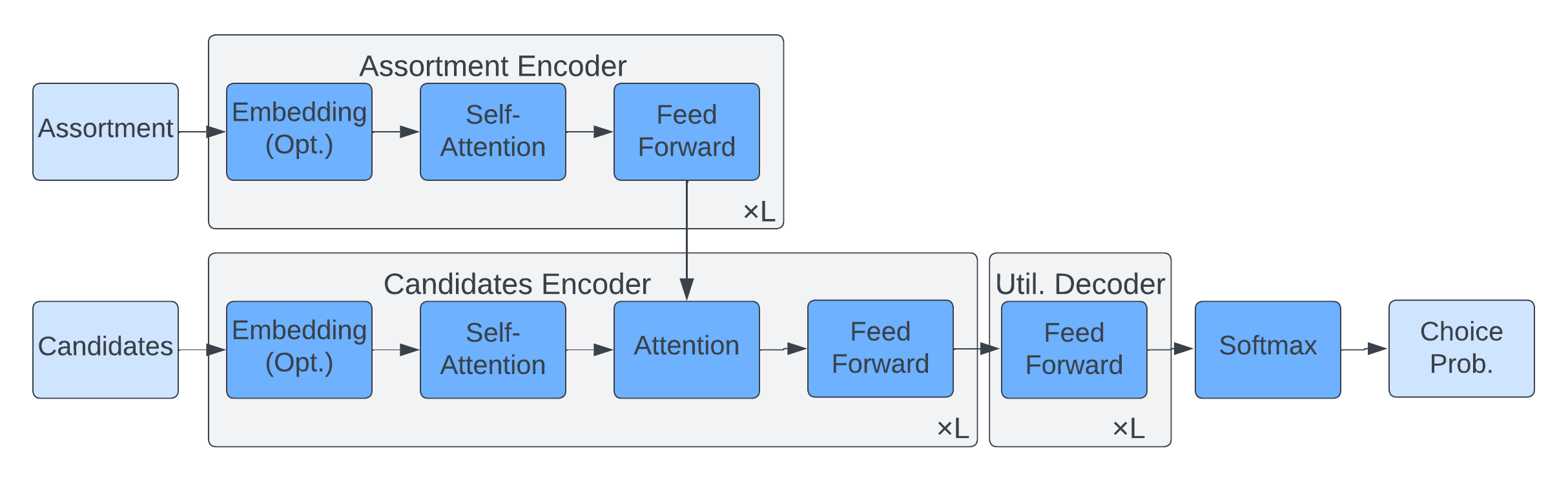}
    \caption{Architecture of Transformer Choice Net (TCNet). We omit the residual connections within each sub-layer.}
    \label{fig:choice_model}
\end{figure*}

Figure \ref{fig:choice_model} illustrates the architecture of the Transformer Choice Net $f^{\text{TCN}}$. The network's input comes from two sources: the assortment set $\mathcal{S}$ of size $S$ and the candidate set $\mathcal{C}$ of size $C$. At noted in \S\ref{sec:setup}, we suppose each item $i$ is associated with a feature vector $x_i\in\mathbb{R}^d$ (either product feature or one-hot encoding of the product id). Then the input matrices of TCN are $X^{\mathcal{S}}\in \mathbb{R}^{S\times d}$ and $X^{\mathcal{C}}\in \mathbb{R}^{C\times d}$, and the output is a vector of choice probabilities for each item $i\in \mathcal{C}$. 

The TCN consists of three main parts: assortment encoder, candidates encoder, and utility decoder.

The \textit{assortment encoder} processes the input $X^{\mathcal{S}}$ and encodes both the original feature $X^{\mathcal{S}}$ and the interaction effects (from the assortment) into a latent feature matrix $\tilde{X}^{\mathcal{S}} \in \mathbb{R}^{S\times d_v}$. Here each item $i$ is associated with a latent feature vector $\tilde{X}^{\mathcal{S}}_i \in \mathbb{R}^{d_v}$. In this feature transformation,  the assortment encoder can contain $L$ layers (which is a hyperparameter), and each with three sub-layers:

First, an optional embedding sub-layer processes the original input $X^{\mathcal{S}}$ by an element-wise fully connected feed-forward network layer applied to each item in the input assortment matrix separately, via two linear operations and an activation function $\phi$:
$$f^{\text{FFN}}(x)\coloneqq W_2\phi(W_1x+b_1)+b_2,$$
where $W_1\in \mathbb{R}^{ d_v\times d}$, $W_2\in \mathbb{R}^{ d_v\times d_v}$ and $b \in \mathbb{R}^{d_v}$ are trainable parameters. The dimension $d_v$ controls the embedding's complexity. 

Second, a self-attention sub-layer implements the attention mechanism defined in \eqref{eqn:self_att}. This layer captures the interaction effects between the items within the assortment. 

Third, a fully connected feed-forward network layer $f^{\text{FFN}}(x)$ (similar to the first part) is applied to each item separately to further process each item's latent features.

The \textit{candidates encoder} processes the original input $X^{\mathcal{C}}$ in a similar manner as the assortment encoder. It processes the original feature and captures the interaction effect between items within the candidate set through three modules as the assortment encoder: an optional embedding sub-layer, a self-attention sub-layer, and a fully-connected feed-forward layer. One special point is an additional cross-attention sub-layer that mingles the assortment set and the candidate set. Specifically, this assortment-candidate attention sub-layer takes (i) the output of the assortment encoder $\tilde{X}^{\mathcal{S}}\in \mathbb{R}^{S\times d_v}$ and (ii) the output $X$ of the self-attention sub-layer of the candidates encoder as input, and then maps them to a latent feature matrix $\tilde{X}^{\mathcal{C}\cdot \mathcal{S}}\in \mathbb{R}^{C\times d_v}$ that is aware of the assortment context. Mathematically, this is represented by
$$f^{\text{AC}}(X)=\text{Att}(XW_Q,\Tilde{X}^{\mathcal{S}}W_K,\Tilde{X}^{\mathcal{S}}W_V)$$
where $W_Q,W_K,W_V\in\mathbb{R}^{d_v\times d_v}$ are trainable parameters.

The \textit{utility decoder} processes the output latent features from the candidates encoder into some latent utilities for each item in $\mathcal{C}$. Ideally, the output latent feature $\tilde{X}^{\mathcal{C}\cdot \mathcal{S}}_i$ from the candidates encoder has encoded both the candidate set effect and the assortment set effect. The utility decoder transforms each of this latent feature vector $\tilde{X}^{\mathcal{C}\cdot \mathcal{S}}_i$ into a (scalar) latent utility $\tilde{u}_i$, and the transformation shares the same weight/parameter across different items. Here we apply a fully-connected feed-forward network, which can be replaced by other network architectures. Lastly, an MNL model/softmax layer converts these latent utilities into choice probabilities.

\subsection{Remarks on TCNet}

We make the following remarks on the TCNet.

\textbf{Assortment size flexibility}. One advantage of the attention mechanism is that it can handle input (e.g., assortment set and candidate set) with varying sizes. This is useful in a choice modeling context because the assortment and the candidate sets may have variable cardinality. Moreover, the number of parameters in TCNet does not depend on the total number of items $|\mathcal{N}|$ which makes it attractive for the case when there are a large number of items or newly added items. As long as the item/product feature of a new item/product is available, the TCNet can be used seamlessly for assortments with newly launched products.

\textbf{Permutation equivariance}. For any permutation $\sigma$ applied to the rows of $ X$ (such as interchanging the first and second row), the $f^{\text{Att}}$ satisfies \textit{Permutation Equivariant} \citep{lee2019set}, which is defined as $ f^{\text{Att}}(\sigma(X)) = \sigma(f^{\text{Att}}(X))$. This ensures that each output choice probability remains invariant to the ordering/position of the input. This is useful for choice modeling context where the items in the assortment/candidate set are presented to the customer all at once. In this case, there is no ordering of the products and thus the customer choice should be permutation equivariant. Of course, when the position/order does matter in determining choice, one can encode the position/order as a feature.

\begin{table}[ht!]
    \centering
    \captionsetup{font=small}
    \resizebox{\columnwidth}{!}{%
    \begin{tabular}{c|c|c|c}
    \toprule
    Model & \# parameters & Variable size & Assortment effect \\ \midrule
    TCNet (Ours) &  $O\left(dd_v+Ld^2_v\right)$ & Yes & Yes  \\
    DeepMNL\citep{wang2020deep, sifringer2020enhancing} &$O\left(dd_v+Ld^2_v\right)$ & Yes &No \\AssortNet \citep{wang2023neural}&$O\left(\|\mathcal{N}\|dd_v+Ld^2_v\right)$&No&Yes\\
    SDANet \citep{rosenfeld2020predicting}&$O\left(dd_v+Ld^2_v\right)$&Yes&Yes
    \\
    FATENet \citep{pfannschmidt2022learning}&$O\left(dd_v+Ld^2_v\right)$&Yes&Yes\\
    DLCL \citep{tomlinson2021learning}&$O\left(d^2\right)$&Yes&Yes\\
RUMNet\citep{aouad2022representing}&$O\left(dd_v+Ld^2_v\right)$&Yes&No\\
     \bottomrule
    \end{tabular}}
    \caption{Comparison on the model complexity, where $d$ and $d_v$ are dimensions of item and latent features  respectively, $L$ is the number of layers, and $\|\mathcal{N}\|$ is the total number of items.}
    \label{tab:complex}
\end{table}

\textbf{Comparison against other NN-based choice models.} In Table 1, we make a comparison between TCNet and several other neural network-based choice models (for single-choice) from several aspects. The column ``variable size'' refers to whether the model requires a fixed assortment size or allows a variable size. The column ``assortment effect'' refers to whether the model captures the assortment effect, i.e., the interactions between items within the assortment. 

In terms of the number of parameters, we can see that TCNet has the same order of the number of parameters as other models, thus its empirical advantage (in the numerical experiment) should be attributed to its architecture rather than model size. In particular, we note that SDANet \citep{rosenfeld2020predicting} and FATENet \citep{pfannschmidt2022learning} essentially implement the Deep Set \citep{zaheer2017deep} for choice modeling, and thus capture item-wise interaction on a coarser scale than our TCNet. DLCL \citep{tomlinson2021learning} explicitly models product interactions through linear functions and does not introduce a neural network architecture.

\textbf{Transformer specializations for choice modeling.} The Transformer \citep{vaswani2017attention} is originally designed for seq2seq tasks, such as machine translation and text summarization. The model predicts the likelihood of the subsequent word from the vocabulary list. In comparison, the TCNet is designed for a set-to-member task, where the goal is to forecast the item(s) to be chosen from the candidate set and thus it requires the permutation equivariance. This necessitates modifications to the input's structure and the utility decoder. Similarly, \citet{lee2019set} introduce the Set Transformer which is designed to model interactions within an input set. Different from TCNet, the Set Transformer aims for permutation invariance ($f(X)=f(\sigma(X))$ for any permutation $\sigma$ over an ordered set $X$) which keeps the whole output invariant through permuting inputs. In the language of choice modeling, permutation invariance assumes that the assortment effect will take place through the assortment set as a whole $\mathcal{S}$ (or $\mathcal{C}$) and keep same for all items, while the permutation equivariance allows a finer interaction between the items.

\textbf{Training:} We use the cross-entropy loss (CE loss) for the training. With dataset $\mathcal{D}=\{(i_k,\{\mathcal{C}_k,\mathcal{S}_k\}), k=1,\ldots,m\}$, the CE loss is defined by
\begin{equation}
\label{eqn:CE_loss}
    L_{\text{CE}}=-\frac{1}{m}\sum_{k=1}^m\log(f^{\text{TCN}}_{i_k}(\mathcal{C}_{k},\mathcal{S}_k)).
\end{equation}
Several tricks/techniques are used in the original Transformer \citep{vaswani2017attention} such as layer normalization, multi-head attention, residual connection, and dropout. We also implement them for TCNet.

\section{Expressiveness of TCNet}
\label{sec:util}
The seminal work \citep{manski1977structure} shows that any single-choice model can be represented by $O(2^n\times n)$  utility parameters: $\{u^{\mathcal{S}}_{i},i \in \mathcal{N}, \mathcal{S}\subseteq \mathcal{N}\}$. Specifically, for all $i\in\mathcal{S} \text{ and } \mathcal{S}\subseteq \mathcal{N}$,
\begin{equation}
\prob(i|\mathcal{S})=\frac{\exp(u^{\mathcal{S}}_{i})}{\sum_{j\in \mathcal{S}}\exp(u^{\mathcal{S}}_{j})}
\label{eqn:prob_util_CM}
\end{equation}
where the utility $u_{i}^\mathcal{S}$ represents the \textit{contextualized} utility of the $i$-th item under the assortment $\mathcal{S}$. \citet{batsell1985new} further show that the utility $u_{i}^\mathcal{S}$ can be decomposed into a sum of item-wise interactions among the subsets of the assortment which is also recently discussed by \citet{seshadri2019discovering}.
\begin{theorem}[\citep{batsell1985new}]
\label{thm:util_decompose}
For any $\{u_{i}^\mathcal{S}: i \in \mathcal{N}, \mathcal{S}\subseteq \mathcal{N}\}$, there exists a unique set of parameters $\{v_{i}^\mathcal{S}: i \in \mathcal{S}, \mathcal{S}\subseteq \mathcal{N}\}$ where $\sum_{i\notin \mathcal{S}}v_{i}^\mathcal{S}= 0, \forall \mathcal{S} \subset \mathcal{N}$, such that for all $i\in\mathcal{S} \text{ and } \mathcal{S}\subset \mathcal{N}$,
$$u_{i}^{\mathcal{S}}=\sum_{\mathcal{S}'\subseteq \mathcal{S}\setminus i}v_{i}^{\mathcal{S}'}.$$
\end{theorem}

The above theorem shows that to estimate a single-choice model, it is enough to estimate the interaction effects of each item $i$, i.e.,
$$\{v_{i}\},\{v_{i}^{\{j\}}: j\in \mathcal{N}, j\neq i\},\ldots, \{v_{i}^{\mathcal{N}\setminus \{i\}}\}.$$
Here the constraints $\sum_{j\notin \mathcal{S}}v_{j}^{\mathcal{S}}= 0, \forall \mathcal{S} \subset \mathcal{N}$ ensure the uniqueness of the parameters (noting the scale-invariance of the utilities in \eqref{eqn:prob_util_CM}). Based on the level of interaction, we call $\{v_{i}\}$ as 0-th order interaction, $\{v_{i}^{\{j\}}\}$ as 1-st order (or pairwise) interaction, and so on. The classic MNL model captures only 0-th order interaction, and several studies \citep{seshadri2019discovering,bower2020preference,yousefi2020choice,tomlinson2021learning,najafi2023assortment} explore diverse formulations to capture 1st-order interactions $\{v_{i}^{\{j\}}\}$. In this sense, TCNet can capture $L$-th order interactions with $L$ the number of layers in the self-attention module. Thus $L$ provides a statistical handle for controlling the model complexity and also a modeling handle for controlling the level of item-wise interactions.

In parallel, one can parameterize any sequential choice model with utility parameters:
 \begin{equation}
\prob(i|\mathcal{C},\mathcal{S})=\frac{\exp(u^{\mathcal{C},\mathcal{S}}_{i})}{\sum_{j\in \mathcal{C}}\exp(u^{\mathcal{C},\mathcal{S}}_{j})},
\label{eqn:prob_util_seq}
\end{equation}
where the utilities can be further decomposed by
$$u_{i}^{\mathcal{C},\mathcal{S}}=\sum_{\mathcal{C}'\subseteq \mathcal{C}\setminus i}v_{i}^{\mathcal{C}',\mathcal{S}}.$$
Here the term $v_{i}^{\mathcal{C}',\mathcal{S}}$ captures the interaction effects of (i) the candidate subset $\mathcal{C}'$ towards item $i$ and (ii) the assortment effect from $\mathcal{S}.$ The interaction from the subset $\mathcal{C}'$ can be viewed as a candidate interaction effect, while the interaction from $\mathcal{S}$ can be viewed as a background contextual effect. This decomposition provides intuition for why the TCNet architecture is suitable for the choice modeling task: the assortment encoder encodes the global contextual effect from $\mathcal{S}$, and the candidate encoder encodes the local candidate interaction effect.

Theoretically, this utility decomposition also provides a guarantee on the model expressiveness of TCNet for the choice modeling task as follows.
\begin{theorem}
\label{thm:capacity}
Any sequential choice model can be represented by a Transformer Choice Net. Specifically, for any sequential choice model defined by \eqref{seq_choice_model}, there exists a Transformer Choice Net $f^{\text{TCN}}$ such that
$$\mathbb{P}(i|\mathcal{C},\mathcal{S})=f^{\text{TCN}}_i(\mathcal{C},\mathcal{S})$$
holds for all $i\in \mathcal{C}, \mathcal{C}\subseteq \mathcal{S}$ and $\mathcal{S}\subseteq \mathcal{N}$.
\end{theorem}
The proof is deferred to Appendix \ref{apx:proof}. It is inspired by \citep{lee2019set} which proves the approximability of the Transformer Set for the permutation invariant function family (but not the permutation equivariant needed here for choice modeling). Combined with our discussion in \S\ref{sec:setup}, we yield the following result.

\begin{corollary}
Any single-choice model or multi-choice model can be represented by a Transformer Choice Net.
\end{corollary}

\section{Numerical Experiments}

In this section, we present numerical experiments to demonstrate the performance of TCNet. Then we give visualizations and interpretations of the learned TCNet.

\subsection{Numerical Performance on Real Datasets}
\label{sec:exps}

We evaluate TCNet's performance against multiple benchmark models for the three choice modeling tasks we consider in this paper: single-choice, sequential choice, and multi-choice. We emphasize that for all the models and datasets, we use a single architecture of TCNet to showcase its power of ``one-size-fits-all''. 

Each experiment is conducted 5 times, with training, validation, and test data randomly split at ratios of $60\%, 20\%$, and $20\%$, respectively. The reported numbers are based on the average values over the test data, accompanied by the standard deviations. Further details on the datasets, the implementation, and the hyperparameter tuning are given in Appendix \ref{apx:exps}.

\textbf{Single-choice prediction.} We use the cross-entropy (CE) loss \eqref{eqn:CE_loss} to train the TCNet and also as the performance metric to report the test performance in single-choice prediction. We compare the TCNet against several benchmark models: DeepMNL \citep{wang2020deep, sifringer2020enhancing}, AssortNet \citep{wang2023neural}, SDANet \citep{rosenfeld2020predicting}, FATENet \citep{pfannschmidt2022learning}, Mixed-MNL \citep{mcfadden2000mixed}, and DLCL \citep{tomlinson2021learning}. These benchmarks represent the state-of-the-art methods of single-choice modeling. Our experiment includes the public datasets used in these existing works: Car \citep{mcfadden2000mixed}, Sushi \citep{kamishima2003nantonac}, Expedia(book), Hotel \citep{bodea2009data}, SFwork and SFshop \citep{koppelman2006self}; as well as two private datasets Flight, and Retail.

\textbf{Sequential choice prediction.} We also use the CE loss as the training objective and the performance metric in sequential choice prediction. We consider two public datasets commonly used in existing works: Expedia(click) and Bakery. These two datasets are originally multi-choice datasets without purchasing sequence information. We transform them into the sequential choice data by a method described in Appendix \ref{apx:exps}. Though several algorithms are developed for sequential choice prediction, their limitations make them inapplicable with these specific datasets. As a result, for benchmarking purposes, we formulate sequential choice prediction as a single-choice prediction problem and utilize the above-mentioned benchmark models. 

Specifically, for a sequential choice dataset denoted as $\mathcal{D}=\{(i_k, \{\mathcal{C}_k,\mathcal{S}_k\}), k=1,...,m\}$, we drop $\mathcal{S}_k$ and recast it as a single-choice dataset: $\mathcal{D}=\{(i_k, \mathcal{C}_k), k=1,...,m\}$ for both the training and testing of the single-choice benchmark models. In essence, these models predict a single-choice from the candidate set \(\mathcal{C}\) as the assortment and omits the original assortment \(\mathcal{S}\), which may cause the information from what have been chosen unavailable (i.e., the assortment context) and only focus on modeling the candidates effect.

To assess the implications of overlooking the assortment context, we train two variants of the Transformer Choice Net: TCNet(Var) and TCNet, where "Var" denotes the variant version. The former is trained as the single-choice model based on the above method while the latter is trained by minimizing the CE loss \eqref{eqn:CE_loss} over the original sequential choice dataset.

\textbf{Multi-choice prediction.}
We adopt the F1 score loss as the performance metric for multiple choices prediction, in line with the approach in \citep{pfannschmidt2022learning}. Given the dataset $\mathcal{D}=\{(\mathcal{B}_k,\mathcal{S}_k), k=1,\ldots,m\}$, the F1 score loss is defined by
\begin{equation*}
      L_{\text{F1}}=1-\frac{2}{m}\sum_{k=1}^m\frac{|\hat{\mathcal{B}}_k\cap\mathcal{B}_k|}{|\hat{\mathcal{B}}_k|+|\mathcal{B}_k|}.
\end{equation*}
Here, $\hat{\mathcal{B}}_k$ represents the algorithm's predicted multiple choices. This loss provides a balanced measure of model performance on imbalanced datasets, which is the case for most choice modeling data (far fewer chosen item(s) than unchosen ones).  

We focus solely on the Expedia dataset here. The Bakery dataset, due to its consistent assortment and unvarying item features, results in identical multi-choice predictions across all samples when deterministic prediction methods are employed which renders it unsuitable for algorithmic comparison. 

As discussed in \S\ref{sec:review}, models designed for multiple choices become computationally expensive with increasing assortment sizes. Following the strategy in \citep{pfannschmidt2022learning}, we adopt several single-choice models for multi-choice prediction: if an algorithm can produce interim utilities, $u_i^{\mathcal{S}}$ for each item $i$ within assortment $\mathcal{S}$, then the item's choice probability is
$\frac{\exp(u_i^{\mathcal{S}})}{1+\exp(u_i^{\mathcal{S}})}.$
And resultant predicted multiple choices are
\begin{equation}
    \left\{i\in\mathcal{S}: \frac{\exp(u_i^{\mathcal{S}})}{1+\exp(u_i^{\mathcal{S}})}>\mu\right\}
    \label{eqn:multi_thrs}
\end{equation}
Here, the threshold $\mu$ is a tunable hyperparameter. In this case, the training loss remains the CE loss, yet is applied to each item's choice outcome independently. We use TCNet to denote the TCNet trained by the method discussed in \S \ref{sec:learning} and use TCNet(Var) as the one trained by the above method.


\begin{table*}[ht!]
\centering
\resizebox{\columnwidth}{!}{%
\begin{tabular}{cc|c|c|c|c|c|c|c|c}
\toprule
Task&Data&TCNet&TCNet(Var)&DeepMNL&AssortNet&SDANet&FATENet&Mixed-MNL&DLCL
\\
\midrule
\multirow{8}{*}{Sin.}&Car&\textbf{1.517}$\pm$0.026 &- &1.578$\pm$0.013 &1.585$\pm$0.032 &\textbf{1.517}$\pm$0.026 &1.521$\pm$0.024 &1.579$\pm$0.029 &\textbf{1.517}$\pm$0.025\\
&Sushi&\textbf{1.747}$\pm$0.011 &- &1.789$\pm$0.007 &1.914$\pm$0.016 &\textbf{1.747}$\pm$0.013 &1.750$\pm$0.010 &1.764$\pm$0.008 &1.763$\pm$0.007\\
&Expedia(book)&\textbf{2.314}$\pm$0.021&- &2.408$\pm$0.014 &2.389$\pm$0.021 &2.328$\pm$0.020 &2.360$\pm$0.019 &2.351$\pm$0.020  &2.340$\pm$0.017\\
&Hotel&\textbf{0.778}$\pm$0.008 &- &1.321$\pm$0.041 &0.851$\pm$0.008 &\textbf{0.778}$\pm$0.006 &0.803$\pm$0.006 &0.882$\pm$0.004&0.885$\pm$0.003\\
&SFwork&\textbf{1.539}$\pm$0.031 &- &1.902$\pm$0.167 &1.667$\pm$0.026&1.542$\pm$0.027 &1.570$\pm$0.033 &1.664$\pm$0.022 &1.683$\pm$0.014\\
&SFshop&\textbf{0.821}$\pm$0.006 &- &1.265$\pm$0.140 &0.928$\pm$0.024 &\textbf{0.821}$\pm$0.010 &0.831$\pm$0.015 &0.909$\pm$0.022 &0.894$\pm$0.011\\
&Flight&\textbf{0.412}$\pm$0.007 &- &2.819$\pm$0.051 &2.751$\pm$0.049&0.427$\pm$0.007 &0.477$\pm$0.006 &0.476$\pm$0.005 &0.477$\pm$0.006\\
&Retail&\textbf{0.982}$\pm$0.021 &- &1.429$\pm$0.073 & 1.032$\pm$0.008&0.992$\pm$0.022 &0.996$\pm$0.023 &0.997$\pm$0.023 &0.997$\pm$0.023\\
\midrule
\multirow{2}{*}{Seq.}&Bakery&2.738$\pm$0.024 &\textbf{2.710}$\pm$0.029 &3.817$\pm$0.006 &2.767$\pm$0.016 &2.792$\pm$0.009 &3.808$\pm$0.006 & 3.557$\pm$0.001&3.811$\pm$0.006\\
&Expedia(click)&\textbf{2.863}$\pm$0.017&2.890$\pm$0.023 &2.913$\pm$0.020 &2.922$\pm$0.015 &3.067$\pm$0.023 &3.074$\pm$0.023 &2.924$\pm$0.024&2.937$\pm$0.012\\
\midrule
Mul.&Expedia(click)&\textbf{0.576}$\pm$0.005&0.583$\pm$0.003&0.659$\pm$0.003 &0.653$\pm$0.005 &0.593$\pm$0.002 &0.639$\pm$0.005&- &- \\
\bottomrule
\end{tabular}}
\caption{Test performances ($\downarrow$) on three tasks: single-choice (Sin.), sequential choice (Seq.), and multi-choice (Mul.).}
\label{tab:test_exp}
\end{table*}

\textbf{Results.} The performances on the test sets across all the experiments are summarized in Table \ref{tab:test_exp}. The Transformer Choice Net (TCNet) exhibits superior performance across all three tasks and datasets. One interesting observation is about the sequential choice prediction on the Bakery data: the TCNet(Var), which is trained without feeding the assortment information outperforms the standard TCNet, which enjoys the assortment information. One explanation is that the uniformity of assortments in all Bakery data samples allows an implicit encoding of this invariant assortment information. Conversely, in the Expedia(click) dataset, where assortments largely vary across samples, the provision of the additional assortment information appears beneficial, enhancing the performance of the standard TCNet. Also, as noted in the last section, the TCNet has a comparable number of parameters as the benchmark networks, so its advantage should be attributed to its explicit modeling of the item interaction.

\subsection{Probing into the Representation of TCNet}
\label{sec:feature_int}

We train a TCNet on the Bakery dataset which contains a total of $50$ items, including food (e.g., cakes with different flavors, and snacks like cookies) and drinks (e.g., different coffees). The original dataset contains 4 item features. However, we train the model without these features and only use a one-hot encoding for each item. We then extract the latent features of each product from the assortment encoder and do linear probing \citep{belinkov2022probing} of the model through running linear logistic regression for two tasks: (1) identify cakes from snacks; (2) identify drinks from food. Due to the limited space, we only show the first here and defer the second to the Appendix.

\begin{figure}[ht!]
  \centering
  \begin{subfigure}[b]{0.43\linewidth}
    \includegraphics[scale=0.45]{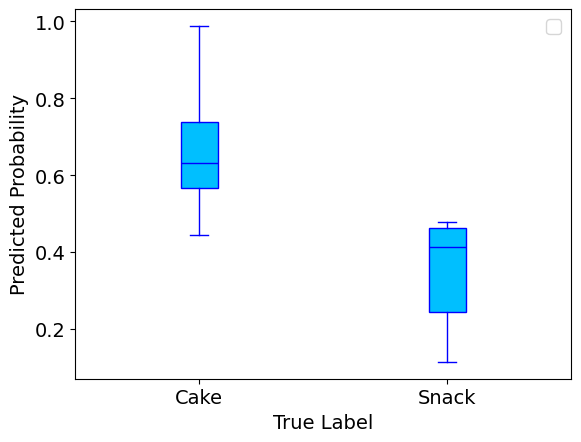}
    \caption{Predicted probability.}
    \label{fig:bplot_cake}
  \end{subfigure}
    \begin{subfigure}[b]{0.43\linewidth}
    \includegraphics[scale=0.43]{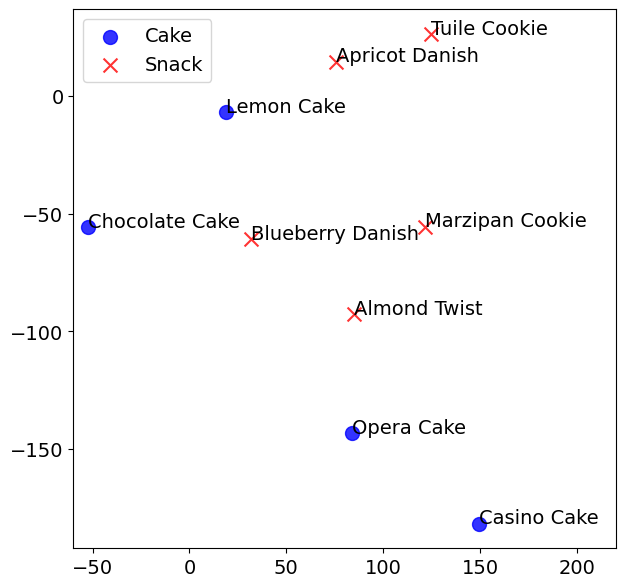}
          \caption{t-SNE visualization of latent features.}
    \label{fig:scat_cake}
  \end{subfigure}
    \caption{Probing the latent representation.}
  \label{fig:feature_vis}
\end{figure}

Figure \ref{fig:feature_vis} showcases our results: Figure \ref{fig:bplot_cake} shows the predicted probabilities to be the cake, and  Figure \ref{fig:scat_cake} uses t-SNE \citep{van2008visualizing} to visualize the top-10 latent features with largest absolute coefficients from the logistic regression. Both figures are based on the testing samples. We note that the latent representation indeed has the ability to identify Cake from Snack, although this is not explicitly included as the feature (only implicitly encoded in the choice). This shows that by learning the choice model from customer behaviors, the latent features can uncover useful properties of items, such as their categories and characteristics.

\subsection{Attention Visualization}
\label{sec:attn_vis}
The attention scores in TCNet across various attention sub-layers reveal the degree of "attention" allocated by the model to each item (whether from an assortment or candidates set) while processing a specific item of interest. Essentially, these scores give insights into the influences and correlations (such as substitutes) of items on the final choice prediction. Moreover, they can help to verify whether the model is concentrating on the important items as we expect. In this subsection, we visualize the attention scores from a trained TCNet as an example to demonstrate how we can understand and explain them.

\begin{figure}[ht!]
  \centering
  \captionsetup{font=small}
  \begin{subfigure}[c]{0.48\linewidth}
    \includegraphics[width=\linewidth]{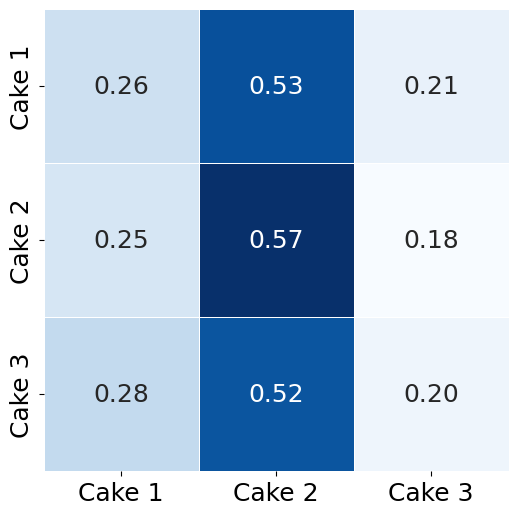}
    \caption{Candidates attention}
    \label{fig:Exp1_1}
  \end{subfigure}
  \hfill 
  \begin{subfigure}[c]{0.48\linewidth}
    \includegraphics[width=\linewidth]{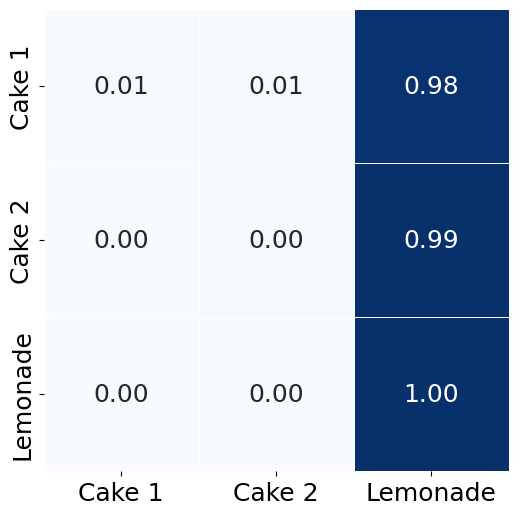}
    \caption{Candidates attention}
    \label{fig:Exp1_2}
    \end{subfigure}
    \begin{subfigure}[c]{0.48\linewidth}
    \includegraphics[width=\linewidth]{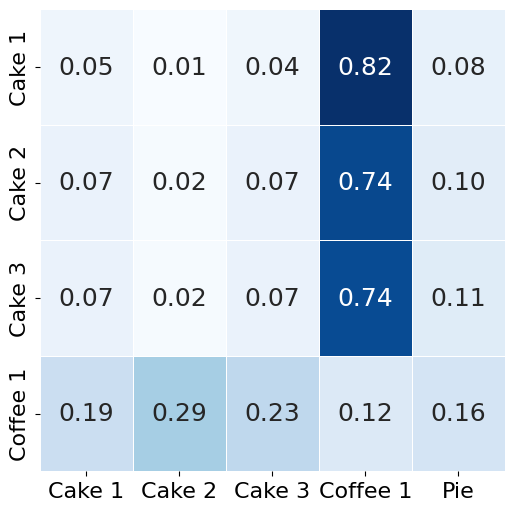}
    \caption{ Cross attention.}
    \label{fig:Exp2_1}
  \end{subfigure}
  \hfill
    \begin{subfigure}[c]{0.48\linewidth}
    \includegraphics[width=\linewidth]{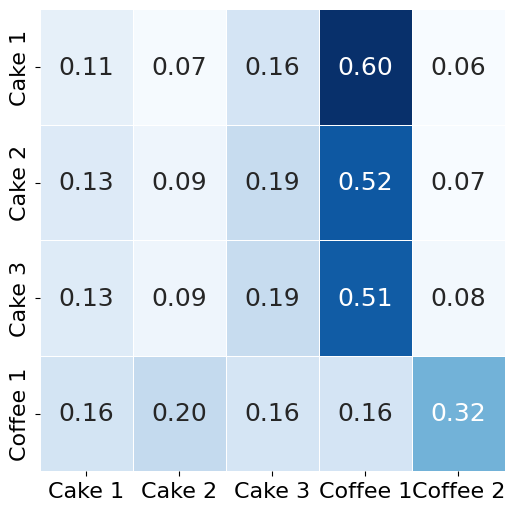}
    \caption{Cross attention.}
    \label{fig:Exp2_2}
  \end{subfigure}
  \caption{Candidates encoder's (self) attention and cross-attention scores. For the second row, the side items are from candidates $\mathcal{C}$ and the bottom items are from assortment $\mathcal{S}$. Each row indicates the normalized attention scores for the side item. }
  \label{fig:attn_exp}
\end{figure}

The TCNet is again trained on the Bakery dataset. Figure \ref{fig:attn_exp} shows two sets of self-attention scores (Figure \ref{fig:Exp1_1} and \ref{fig:Exp1_2}) and two cross-attention scores (Figure \ref{fig:Exp2_1} and \ref{fig:Exp2_2}) from the candidates encoders. We make the following observations:

Comparing Figures \ref{fig:Exp1_1} and \ref{fig:Exp1_2}, we can see Cake 1 and Cake 2 put extremely high weights on the lemonade in Figure  \ref{fig:Exp1_2} in contrast to the Cake 3 offered in Figure \ref{fig:Exp1_1}. This aligns with one's intuition when making the choice: Lemonade, as a drink, represents a new category differing from the other candidates and can impose a substantial influence on the final choice. For instance, a thirsty customer might randomly choose one cake in Figure \ref{fig:Exp1_1} but will definitely choose the lemonade when it is offered as in Figure \ref{fig:Exp1_2}.

In Figures \ref{fig:Exp2_1} and \ref{fig:Exp2_2}, the first observation is consistent with the above: all three cakes assign substantial weights to Coffee 1, which is a drink, in both figures.  Another observation is on the weight of Coffee 1: its weight on Coffee 2 in Figure \ref{fig:Exp2_1} is double of that on the Pie in Figure \ref{fig:Exp2_2}. Given that both Coffee 2 and Pie are purchased items, this implies that buying Coffee 2 has a greater influence on choosing Coffee 1 than purchasing the Pie. This is reasonable, as consumers typically do not buy two coffees but may purchase a coffee alongside a pie.

In summary, examining the attention scores enables us to validate whether the model adheres to our common sense and allows us to get insights into the relationships between items.

\section{Conclusion and Future Directions}
\label{sec:conclu}
The existing literature on single-choice model, sequential choice model, and multi-choice model have been largely separate from each other: the current sequential choice models have there own limitations, and the current multi-choice models become impractical when dealing with a large assortment. 

While the research on single-choice models is vast, the literature on predicting sequential and multi-choice is noticeably less developed. Our paper introduces the Transformer Choice Net as a unified neural network framework capable of addressing all three types of choice behaviors. Notably, it is the first Transformer-based architecture tailored for choice(s) prediction tasks. We prove its universal representation capacity and, through extensive numerical experiments, demonstrate its superior performance over several benchmarks. We conclude with the following future directions.

\textbf{Assortment optimization.} One important downstream application of choice prediction is assortment optimization, which optimizes the offered assortment to maximize the expected revenue for the seller. Given the Transformer Choice Net's strong prediction performance, an intriguing avenue to explore would be the optimization of assortment offerings for a trained Transformer Choice Net model.

\textbf{Exploring alternative Transformer-based structures.} The Transformer architecture has increasingly shown its significance within the deep learning realm, showcasing its adaptability across diverse tasks. While the Transformer Choice Net represents an initial endeavor in integrating the Transformer framework into choice modeling, we believe there is still large potential in exploring other Transformer-inspired architectures.

\bibliographystyle{informs2014}
\bibliography{main}

\appendix
\section{Proof for Theorem \ref{thm:capacity}}
\label{apx:proof}
\begin{proof}
For items $\mathcal{N}=\{1,2,\ldots,n\}$, assume the underlying sequential choice model is represented by parameters $\{u_{i}^{\mathcal{C},\mathcal{S}}\}$ for \eqref{eqn:prob_util_seq}. Then it is sufficient to prove the output virtual utilities of some TCNet (with specific parameters) $\Tilde{u}_{i}^{\mathcal{C},\mathcal{S}}=u_{i}^{\mathcal{C},\mathcal{S}}$ for all $i,\mathcal{C},\mathcal{S}$. Since the items are finite, without loss of generality,  we assume the input feature of each item $i$ is the one-hot vector with dimension $n$ where only $i$-th element is $1$ and all others are $0$'s.

We first focus on the assortment encoder part. We construct a one-layer assortment encoder without the embedding sub-layer. For the self-attention sub-layer, we set its query weight matrix $W_Q$ as zero matrix, value weight matrix $W_V$ as identity matrix, activation function $\phi(x)=1+\text{Relu}(x)$ and without any residual connection. Under the above setting, the self-attention sub-layer just sums all rows in the input matrix $X^{\mathcal{S}}$ for each item: for any item $i\in\mathcal{S}$, its output is a vector with dimension $n$: $\sum_{j\in \mathcal{S}}x_j$. Then for the following feed-forward sub-layer, we set all parameters as $0$ to "mute" this sub-layer and set the residual connection to make sure the final outputs of the assortment encoder $\tilde{X}^\mathcal{S}$ has each row  $\Tilde{x}_i^\mathcal{S}=\sum_{j\in \mathcal{S}}x_j$. Indeed, this output uses  $0$ and $1$ to encode the assortment information: the positions with value $1$ indicate the items in the assortment $\mathcal{S}$, and all other positions are the items not in the assortment.

Then for the candidates encoder part, we also construct it with only one layer and without the embedding sub-layer. For its self-attention sub-layer, we use the same structure as in the assortment encoder's self-attention except that we add the residual connection. Thus, with residual connection, the self-attention's output for item $i$ is $x_i+\sum_{j\in \mathcal{C}}x_j$. For the following assortment-candidates attention sub-layer, we again set its query weight matrix $W_Q$ as zero matrix, value weight matrix $W_V$ as identity matrix, with residual connection, but the activation function is now Softmax. Thus, this attention itself (without residual connection) will output the mean of all rows in $\tilde{X}^{\mathcal{S}}$, i.e., $\sum_{j\in \mathcal{S}}x_j$, for each item $i\in \mathcal{C}$. With residual connection, the final output for each item $i$ of this sub-layer is just  $x_i+\sum_{j\in \mathcal{C}}x_j+\sum_{j\in \mathcal{S}}x_j$. We should notice this output vector uses four values to encode all input information: the position with value $3$ indicates the item $i$ we focus on, value $2$'s indicate the items as candidates by noting $\mathcal{C}\subseteq \mathcal{S}$, value $1$'s indicate the items are not candidates but were from the assortment (i.e., items in the basket), and $0$'s indicate the items not in the assortment. Finally, as in the assortment encoder part, for the following feed-forward network, we set all parameters as $0$ to "mute" this sub-layer and set the residual connection to make sure the final outputs of the candidates encoder are still $\tilde{x}_{i}^{\mathcal{C}\cdot \mathcal{S}}=x_i+\sum_{j\in \mathcal{C}}x_j+\sum_{j\in \mathcal{S}}x_j.$

Now we only need to show that the utility decoder can perfectly decode the input information
$\tilde{x}_{i}^{\mathcal{C}\cdot \mathcal{S}}$
to approximate $u_{i}^{\mathcal{C},\mathcal{S}}$. We utilize the universal approximation theorem \citep{hornik1991approximation} to prove it.

We set the utility decoder with one layer:
\begin{multline*}
    \text{decoder}(x)=\sum_{\mathcal{S}\subseteq\mathcal{N},\mathcal{C}\subseteq\mathcal{S},i\in\mathcal{C}}u^{\mathcal{C},\mathcal{S}}_{i}\Bigg( \Bigg.  \text{Relu}(w^\top (x-\tilde{x}_{i}^{\mathcal{C} \cdot \mathcal{S}})-1)+ \text{Relu}(w^\top (x-\tilde{x}_{i}^{\mathcal{C} \cdot \mathcal{S}})+1)-\\ 2\times  \text{Relu}(w^\top (x-\tilde{x}_{i}^{\mathcal{C} \cdot \mathcal{S}}))   \Bigg. \Bigg),
\end{multline*}

where $w=(4^1,4^2,...,4^n)\in\mathbb{R}^n$. Thus, the linear operation $w^\top \tilde{x}_{i}^{\mathcal{C} \cdot \mathcal{S}}$  maps the input information $(i,\mathcal{C},\mathcal{S})$ uniquely to an integer. And by noticing the operation
$$ \text{Relu}(w^\top (x-\tilde{x}_{i}^{\mathcal{C} \cdot \mathcal{S}})-1)+ \text{Relu}(w^\top (x-\tilde{x}_{i}^{\mathcal{C} \cdot \mathcal{S}})+1)-2\times  \text{Relu}(w^\top (x-\tilde{x}_{i}^{\mathcal{C} \cdot \mathcal{S}}))  $$
equals to $1$ when $x=\tilde{x}_{i}^{\mathcal{C}\cdot \mathcal{S}}$ and $0$ for all  $x=\tilde{x}_{i'}^{\mathcal{C}'\cdot \mathcal{S}'}$ where $(i',\mathcal{C}',\mathcal{S}')\neq (i,\mathcal{C},\mathcal{S})$, we can conclude that
$\tilde{u}_{i}^{\mathcal{C},\mathcal{S}}=\text{decoder}(\tilde{x}_{i}^{\mathcal{C}\cdot \mathcal{S}})=u_{i}^{\mathcal{C},\mathcal{S}}.$
\end{proof}

\section{Appendix for Section \ref{sec:exps}}
\label{apx:exps}
This appendix provides detailed information about the datasets and their preprocessing steps, model implementations, training processes, and hyperparameter tuning methods referenced in \S\ref{sec:exps}.

\subsection{Dataset}

\begin{table*}[ht!]
    \centering
    \resizebox{\columnwidth}{!}{%
    \begin{tabular}{c|cccccccc|cc}
    \toprule
&\multicolumn{8}{|c|}{Single Choice}&\multicolumn{2}{c}{Multiple Choice}\\
\midrule
&Car&Sushi&Expedia(book)&Hotel&SFwork&SFShop&Flight&Retail&Bakery&Expedia(click) \\
\midrule
\# Samples&4,652&5,000&40,000&9,330&5,029&3,157&40,000&17,200 &20,000&8,041\\
\# Features&21&7&13&0&0&0&0&0&4&13\\
\midrule
Largest Assortment Size&6&10&39&17&6&8&26&21&50&38\\
Mean Assortment Size&6&10&25.90&10.99&4.38&7.20&10.58&17.20&50&26.85\\
\midrule
Largest Chosen Subset Size&1&1&1&1&1&1&1 &1&8& 10\\
Mean Chosen Subset Size&1&1&1&1&1&1&1 &1&3.55& 4.95\\
 \bottomrule
    \end{tabular}}
    \caption{Summary statistics of datasets.}
    \label{tab:data}
\end{table*}
Table \ref{tab:data} summarizes the datasets used in the experiments and the details are as follows:
\begin{itemize}
    \item Car: The data is avaliable at \url{http://qed.econ.queensu.ca/jae/2000-v15.5/mcfadden-train/}. Each observation contains the single choice from six cars with varying features.  We use $21$ item features: \textit{Prices divided by $\log(income)$, Range, Acceleration, Top speed, Pollution, Size, "Big enough", Luggage space, Operating cost, Station availability, Sports utility vehicle (SUV), Sports car, Station wagon, Truck, Van, Electric Vehicle (EV), Commute < 5 times for EV, College times EV, Constant for Compressed Natural Gas (CNG) ,Constant for methanol, College times methanol}. The details of the features can be found in \citep{mcfadden2000mixed}.
    \item Sushi: The data is avaliable at \url{https://www.kamishima.net/sushi/}. We use the \textit{sushi3b.5000.10.order} dataset, which shows the top-10 ranked sushis for each customer. We use these top-10 sushis as the assortment and the top-1 as the final choice. We use $7$ item features \textit{Style, Major-group, Minor-group, Heaviness, Frequency-eaten, Normalized-price, Frequency-sold}, and $6$ customer features \textit{Gender, Age, Time to fill the survey, Most Long-Lived East/West ID until Age 15, Current East/West ID, Match of Long-Lived and Current Region}. The details of the features can be found in \citep{kamishima2003nantonac}.
    \item Hotel: The data is avaliable at \url{https://pubsonline.informs.org/doi/abs/10.1287/msom.1080.0231} and  from \cite{bodea2009data}. It is originally collected from five U.S. properties of a major hotel chain. Each observation contains different room types to be chosen and the final choice. We use the  Hotel 1's dataset.  We preprocess the data as in \cite{csimcsek2018expectation},  removing purchases without matched room type in the assortment,  room types with few purchases (less than $10$), and adding an auxiliary no-purchase option (create four no-purchase records for each purchase record). 
    \item SFwork and SFshop: The raw data is available at \url{https://github.com/tfresource/modechoice} and originally from \cite{koppelman2006self}. We use the cleaned version from \url{https://github.com/arjunsesh/cdm-icml}. The datasets contain the choices of transportation to commute to shop or work around the San Francisco. Dataset features are not used in our study.
    \item Flight and Retail: These are two private real datasets without features. The airline data contains the offered choices of offered bundles, i.e., bundle of seat, food etc. The offered assortments are decided by both the selling strategy and remaining seats. The retailing data is collected from a  fashion-retailing company, where the items are clothes (from a same category and thus most customers  purchase at most one item) and the assortments are decided by the inventories. Both datasets  contain  records of purchase transactions only, so we add no-purchases following the same way as Hotel dataset.
    \item Bakery: the data is available at \url{http://users.csc.calpoly.edu/~dekhtyar/466-Spring2018/labs/lab01.html}. All observations share the same offered assortment containing $50$ items from a bakery and the final (multiple) chosen items. We also encode each item with a unique one-hot encoder as the feature. We use the \textit{20000} folder's dataset with $4$ item features: \textit{Flavor, Food Type, Price, Food or Drink}. To construct the sequential choice dataset, we randomly sample one item from the chosen subset as the final choice, and add other unchosen items as the candidates (and the assortment are still all $50$ items).
    \item Expedia:  the data is avaliable at \url{https://www.kaggle.com/c/expedia-hotel-recommendations/overview}. It contains the ordered list of hotels according to the user's search, and the clicked and final booked (if any) hotel of each search session. We clean the data by dropping the searches with outlier prices ($>\$1,000$) and days of booking in advance (gap between booking date and check-in date $>365$) and also the features with missing values. Weuse $7$ item features \textit{Hotel star rating, Hotel chain or not, Location score (by Expedia), Historical prices, Price, Promotion or not, Randomly ranked or not (by Expedia)}  and $6$ customer features \textit{\# nights stay, \# days booking in advance,\# adults, \# children,\# rooms,Saturday included or not }. We also use the rankings (in the website) of each item as a feature. To construct the single choice dataset, we choose the booked hotel as the final single choice, and add the no-purchase option if there is no booking. For the multiple choice dataset, we use the sessions with total clicks in the range of $2$ to $10$, and use clicked hotels as the final choices. For the sequential choice dataset we use the same construction method as in the Bakery dataset.
\end{itemize}
In addition, due to the computation resource limitation, we truncate all datasets (Expedia and Flight) with more than $40000$ samples to $40000$ samples.
\subsection{Models, Hyperparameters, and Training}
\begin{itemize}
    \item TCNet: Our implementation of the Transformer Choice Net contains a single-layer candidates encoder and an assortment encoder with one embedding sublayer. The utility decoder also contains a single layer but without an activation function. We employ multi-head attention, layer normalization, residual connection, and dropout as described in \citep{vaswani2017attention}, with the number of heads being a tunable hyperparameter and dropout rate $0.1$. When inferring multi-choice model, we add a virtual stopping item during the training as discussed in \S\ref{sec:learning}.
  \item DeepMNL: We modify the architecture by \cite{wang2020deep,sifringer2020enhancing}. We use a neural network $g$ to model the (mean) utility of each item $i$ as a function of its features $x_i$: $u_i^{\mathcal{S}}=u_i=g(x_i)$ for all $i\in \mathcal{N}$ and $\mathcal{S}\subseteq \mathcal{N}$. The architecture we implement has two layers for the NN. This design is specific to scenarios with observed features; otherwise, it reduces to the MNL model.
  \item AssortNet: We implement the AssortNet architecture by \cite{wang2023neural}. In the absence of features, AssortNet uses a neural network $g$ as an assortment encoder, which directly maps the assortment $\mathcal{S}$ into the assortment-specific utilities. Specifically, for any assortment $\mathcal{S}$, the output $g(\mathcal{S})\in \mathbb{R}^{n}$ models  $u_i^{\mathcal{S}}=g_i(\mathcal{S})$ for $i\in \mathcal{S}$ and $u_i^{\mathcal{S}}=-\infty$ for $i\notin  \mathcal{S}$. When features are present, feature encoders are employed to embed these features. The architectural implementation comprises a single layer for the assortment encoder and two layers for the feature encoders (if needed).
  \item SDANet: We implement the Set-Dependent Aggregation Net (SDANet) by \cite{rosenfeld2020predicting}, which uses two neural networks $w$ and $\phi$ to model the assortment-specific utilities:
$u_i^{\mathcal{S}}=w(\mathcal{S})\cdot \phi(x_i,\mathcal{S}).$
    The exact implemented architecture follows the default setting of the posted code. \url{https://drive.google.com/file/d/1KZVbqfVR6QNIpv38y4e8ptzVdbi_GQH4/view}.
   \item FATENet: We implement the First Aggregate Then Evaluate Net (FATENeT) by \cite{pfannschmidt2022learning}, which uses two neural networks $U'$ and $\phi$ to model the assortment-specific utilities:
$u_i^{\mathcal{S}}=U'\left(x_i,\frac{1}{S}\sum_{j\in\mathcal{S}}\phi(x_j)\right).$
    The exact implemented architecture follows the default setting of their posted code. \url{https://github.com/kiudee/cs-ranking}.
   \item Mixed-MNL: We train a featured mixed-MNL model \citep{mcfadden2000mixed} by setting the number of customer types the same as the number of item features, and each customer type's utility on item $i$ is a linear function of its features. Th
 is follows the setting in \citep{tomlinson2021learning}.
  \item DLCL: We implement decomposed linear context logit (DLCL) by \cite{tomlinson2021learning}, which is an extension of Mixed-MNL such that each customer type's utility on item $i$ is a linear function of both the item $i$ itself and the averaged features over all items in the assortment. The number of customer types is still set by the number of item features, which follows the setting in \citep{tomlinson2021learning}.
\end{itemize}
\subsection{Training Process and Hyperparameters Tuning}
\label{apx:train_process}
We train all models using the Adam optimizer \citep{kingma2014adam}, targeting the minimization of training loss specified in \S\ref{sec:exps}. The learning rate decays exponentially at a rate of 0.95 every 10 epochs. The initial learning rate is tuned from the values $0.001, 0.0005$, and $0.0001$. All trainable parameters are initialized using the uniform Xavier method \citep{glorot2010understanding}. Training has a total of 100 epochs without early stopping, utilizing a mini-batch size of 256. The model with the lowest validation error over the 100 epochs is chosen for testing. Additionally, for models  DeepMNL, AssortNet, SDANet, FATENet, and TCNet, we tune the width of the hidden layers from 32, 128, and 256. For the number of heads in TCNet, we tune it from 4, 8, 16, and 32. For the DeepMNL, AssortNet, SDANet, FATENet, and TCNet in multiple choices prediction task, we tune the threshold defined in \eqref{eqn:multi_thrs} from  0.1, 0.3, 0.5, 0.7, 0.9.

\section{Appendix for Section \ref{sec:feature_int}}
\label{apx:feat_int}

\begin{figure}[ht!]
  \centering
  \begin{subfigure}[b]{0.43\linewidth}
    \includegraphics[scale=0.45]{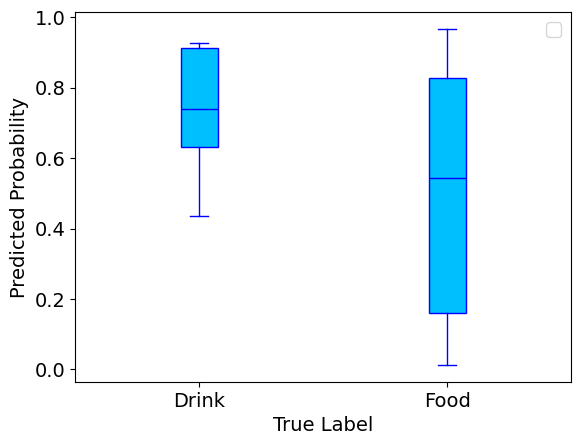}
    \caption{Predicted probability}
    \label{fig:bplot_drink}
  \end{subfigure}
    \begin{subfigure}[b]{0.43\linewidth}
    \includegraphics[scale=0.43]{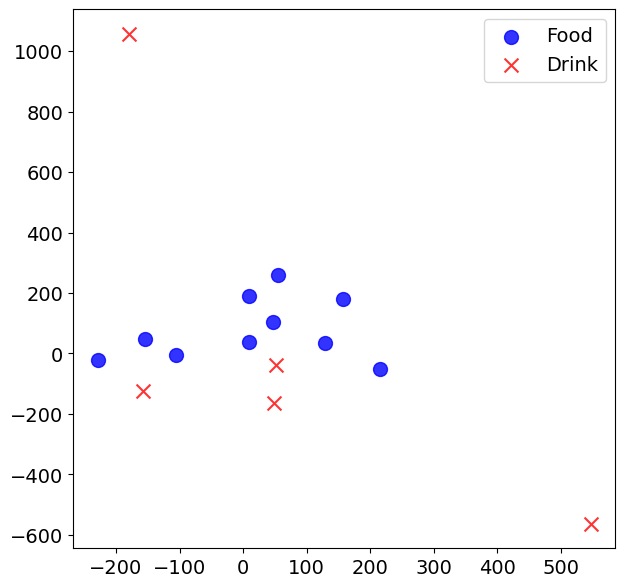}
  \caption{t-SNE visualization of latent features.}
    \label{fig:scat_drink}
  \end{subfigure}
  \caption{Probing the latent representation to identify drink from food.}
  \label{fig:feature_vis_2}
\end{figure}

\textbf{Experiment details}. We train the TCNet with 4 heads on all attention sub-layers, and we choose the initial learning rate as 0.0005 and the hidden dimension as 64. All other hyperparameters, training process, and dataset details are the same as in Appendix \ref{apx:exps}. For training the linear logistic model in Figure \ref{fig:feature_vis}, we select the cakes Strawberry Cake, Truffle Cake, Napoleon Cake, and Apple Pie (not cake but still more expensive and larger compared to snacks), and the snacks Ganache Cookie, Gongolais Cookie, Raspberry Cookie, and Lemon Cookie as training set. The testing dataset for both Figures \ref{fig:bplot_cake} and \ref{fig:scat_cake} are shown in Figure \ref{fig:scat_cake}. For training the logistic model in Figure \ref{fig:feature_vis_2}, we select the items with indices from $35$ to $44$ as the training set, where the first 5 items are food and the others are drinks. The testing dataset for Figure \ref{fig:bplot_drink} contains all other samples, which includes $35$ types of food and $5$ drinks, and testing dataset for Figure \ref{fig:scat_drink} contains the other $5$ drinks and the food items with index from $0$ to $10$ (items information can be found in \url{http://users.csc.calpoly.edu/~dekhtyar/466-Spring2018/labs/lab01.html}).

\section{Appendix for Section \ref{sec:attn_vis}}
\label{apx:vis}
\subsection{Experiment Details}
We train the TCNet with a single head on all attention sub-layers, and we choose the initial learning rate as 0.001 and the hidden dimension as 32. All other hyperparameters, training process, and dataset details are the same as in Appendix \ref{apx:exps}.

\subsection{More Visualization Examples}

We use two more examples to showcase how the attention scores in each attention sub-layer help TCNet encode important items.

We construct a sequential choice model for four items $\mathcal{N}=\{A,A',B,L\}$, where $B$ indicates a benchmark item and $L$ indicates the 'leave' option. We assume items $A'$, $B$ and $L$ have constant utility $1$ independent of $\mathcal{C}$ and $\mathcal{S}$. However, the utility of $A$ can be boosted as follows:

In the first example, the utility of $A$ equals to $100$ if $A'\in \mathcal{C}$ and otherwise equals $1$. This indicates an (extreme) case that the appearance of $A'$ as a candidate can boost the utility of $A$. A similar choice behavior can be found in the experiment by \citep{ariely2008predictably}.
 
In second example, the utility of $A$ equals to $100$ if $A'\in \mathcal{S} \setminus \mathcal{C}$ (and otherwise equals to $1$). This indicates an (extreme) case that the utility of $A$ will be boosted when $A'$ has been chosen. This describes the ``buy one ($A'$) and get another ($A$) free'' situation.

\begin{figure}[ht!]
  \centering
  \captionsetup{font=small}
  \begin{subfigure}[c]{0.48\linewidth}
    \includegraphics[width=\linewidth]{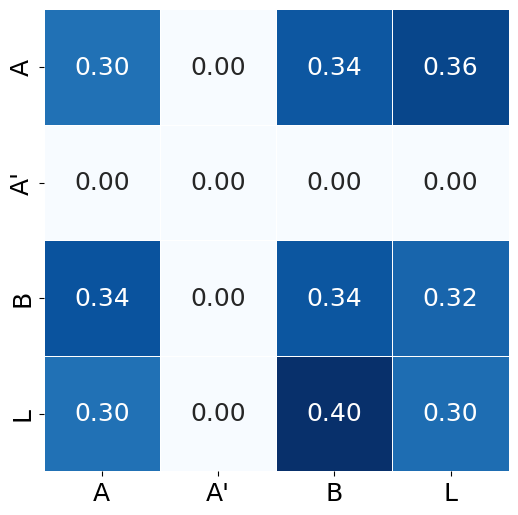}
    \caption{Candidates attention of $(\mathcal{S}_1,\mathcal{C}_1)$.}
    \label{fig:D1_cand_enc_1011}
  \end{subfigure}
  \hfill 
  \begin{subfigure}[c]{0.48\linewidth}
    \includegraphics[width=\linewidth]{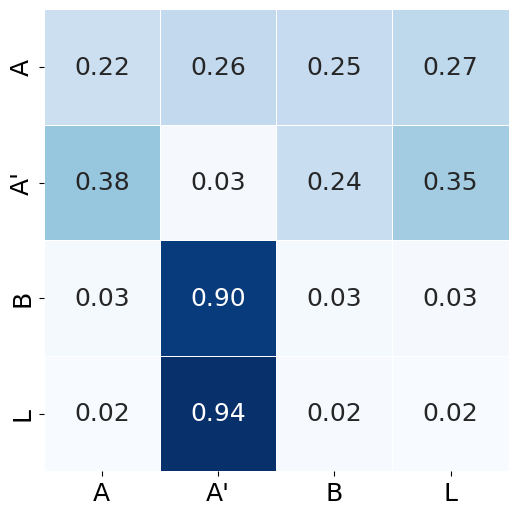}
    \caption{Candidates attention of $(\mathcal{S}_2,\mathcal{C}_2)$.}
    \label{fig:D1_cand_enc_1111}
    \end{subfigure}
    \begin{subfigure}[c]{0.48\linewidth}
    \includegraphics[width=\linewidth]{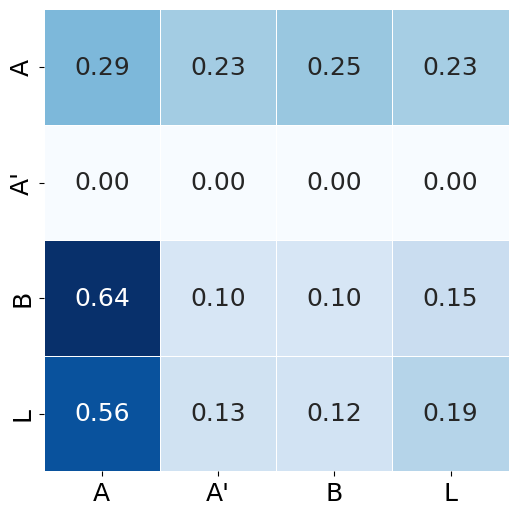}
    \caption{ Cross attention of $(\mathcal{S}_1,\mathcal{C}_1)$.}
    \label{fig:D1_cross_1011}
  \end{subfigure}
  \hfill
    \begin{subfigure}[c]{0.48\linewidth}
    \includegraphics[width=\linewidth]{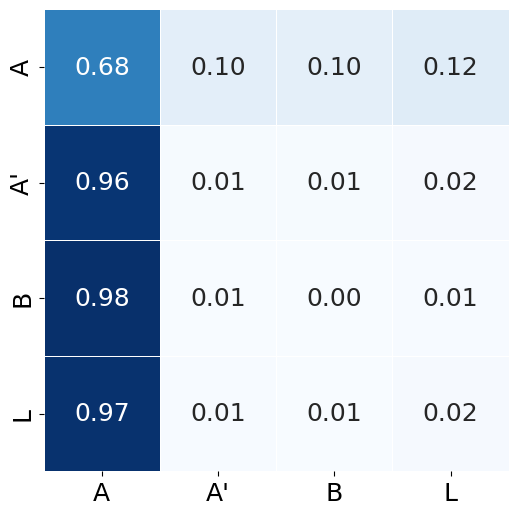}
    \caption{Cross attention of $(\mathcal{S}_2,\mathcal{C}_2)$.}
    \label{fig:D1_cross_1111}
  \end{subfigure}
  \caption{First example: candidates encoder's (self) attention and  cross attention scores for two inputs $(\mathcal{S}_1,\mathcal{C}_1)$ and $(\mathcal{S}_2,\mathcal{C}_2)$, where $\mathcal{S}_1=\mathcal{S}_2=\{A,A',B,L\}$, $\mathcal{C}_1=\{A,B,L\}$, $\mathcal{C}_2=\{A,A',B,L\}$. Each row indicates the normalized attention scores for the labeled item. If the row's corresponding item is not available, we set its attention scores as $0$.  }
  \label{fig:D1_attn}
\end{figure}

\begin{figure}[ht!]
  \centering
  \captionsetup{font=small}
  \begin{subfigure}[c]{0.48\linewidth}
    \includegraphics[width=\linewidth]{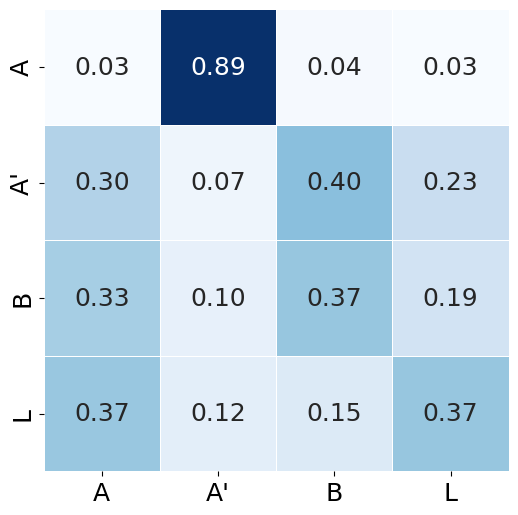}
    \caption{Assortment attention of $(\mathcal{S}_1,\mathcal{C}_1)$.}
    \label{fig:D2_assort_enc_1011}
  \end{subfigure}
  \hfill 
  \begin{subfigure}[c]{0.48\linewidth}
    \includegraphics[width=\linewidth]{figures/D2_assort_enc.png}
    \caption{Assortment attention of $(\mathcal{S}_2,\mathcal{C}_2)$.}
    \label{fig:D2_assort_enc_1111}
    \end{subfigure}
    \begin{subfigure}[c]{0.48\linewidth}
    \includegraphics[width=\linewidth]{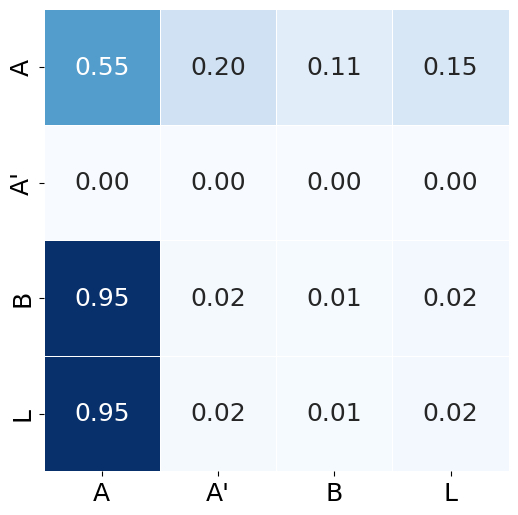}
    \caption{ Cross attention of $(\mathcal{S}_1,\mathcal{C}_1)$.}
    \label{fig:D2_cross_1011}
  \end{subfigure}
  \hfill
    \begin{subfigure}[c]{0.48\linewidth}
    \includegraphics[width=\linewidth]{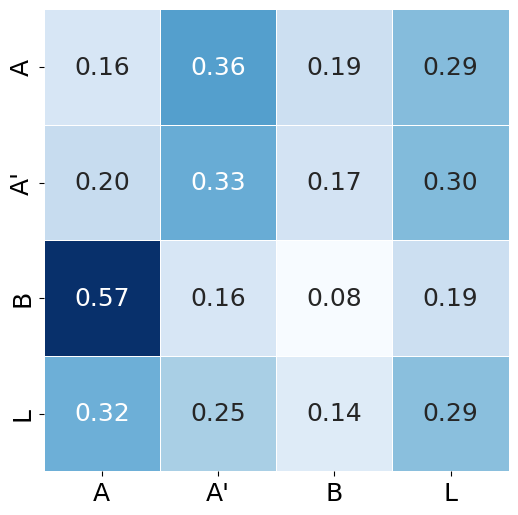}
    \caption{Cross attention of $(\mathcal{S}_2,\mathcal{C}_2)$.}
    \label{fig:D2_cross_1111}
  \end{subfigure}
  \caption{Second example: assortment encoder's self-attention and candidates encoder's cross attention scores for two inputs $(\mathcal{S}_1,\mathcal{C}_1)$ and $(\mathcal{S}_2,\mathcal{C}_2)$, where $\mathcal{S}_1=\mathcal{S}_2=\{A,A',B,L\}$, $\mathcal{C}_1=\{A,B,L\}$, $\mathcal{C}_2=\{A,A',B,L\}$. Each row indicates the normalized attention scores for the labeled item. If the row's corresponding item is not available, we set its attention scores as $0$.  }
  \label{fig:D2_attn}
\end{figure}

We train two TCNets based on these two choice models. Figure \ref{fig:D1_attn}  and \ref{fig:D2_attn} output the attention scores of two examples respectively.
 
Example 1:
Figure \ref{fig:D1_attn} illustrates the (normalized) attention scores derived from the trained TCNet for two inputs: $(\mathcal{S}_1,\mathcal{C}_1)$ and $(\mathcal{S}_2,\mathcal{C}_2)$, where $\mathcal{S}_1=\mathcal{S}_2=\{A,A',B,L\}$, $\mathcal{C}_1=\{A,B,L\}$, $\mathcal{C}_2=\{A,A',B,L\}$. Recall that the attention scores can basically quantify the influence of one item on another.   Figure \ref{fig:D1_cand_enc_1011} presents the candidates encoder's self-attention score matrix, displaying relatively uniform attention allocated to all three candidates in $\mathcal{C}_1$; In contrast, Figure \ref{fig:D1_cand_enc_1111} exhibits a significant increase in attention towards
 $A$ for both $B$ and $L$ when $A'$ is included among the candidates. This shift can reveal $A'\in\mathcal{C}_2$  significantly alters the choice probabilities, unlike in the first scenario where the absence of $A'$ results in fairly balanced attention among the remaining candidates. Figures \ref{fig:D1_cross_1011} and \ref{fig:D1_cross_1111} represent the candidates encoder's cross attention score matrix. There is a noticeable rise in weights on $A$ for all three items $A',B,L$ as depicted in Figure \ref{fig:D1_cross_1111}, compared to Figure \ref{fig:D1_cross_1011}.  This elevation in weights demonstrates the substantial impact $A$ has on the choice probability, attributed to its high utility value.

 Example 2: Figure \ref{fig:D2_attn} shows the attention scores derived from the TCNet again for two inputs same as in Example 1: $(\mathcal{S}_1,\mathcal{C}_1)$ and $(\mathcal{S}_2,\mathcal{C}_2)$. Figure \ref{fig:D2_assort_enc_1011} and Figure \ref{fig:D2_assort_enc_1111} display the assortment encoder's self-attention score matrix for two inputs, which are identical since $\mathcal{S}_1=\mathcal{S}_2$. One observation is the large attention on $A'$ for $A$, which reveals the potentially large influence of $A'$ since it appears in the assortment (while the assortment encoder is currently unknown whether $A'$ is already chosen or not). Figures \ref{fig:D2_cross_1011} and \ref{fig:D2_cross_1111} represent the candidates encoder's cross-attention score matrix. There are extreme weights on $A$ for  $B,L$ as depicted in Figure \ref{fig:D2_cross_1011}, compared to Figure \ref{fig:D2_cross_1111}.  This indicates that the large influence of $A$ on the choice probability, due to the fact that $A'\in\mathcal{S}_1\setminus \mathcal{C}_1$, is encoded in \ref{fig:D2_cross_1011}; While when $A'\notin\mathcal{S}_1\setminus \mathcal{C}_1$, because the (true) choice probabilities are equal among all candidates in Figure \ref{fig:D2_cross_1111}, the weights are more balanced.

Below are the experiment details:

Training data generation: We generate $24,000$ samples for both training datasets, where all samples' assortments are independently and identically sampled by including each item $A,A',B$ with probability $0.5$ and always including $L$. The candidates are similarly sampled by including the items in assortments with probability $0.5$ while always including $L$. Conditional on the sampled assortment and candidates, the final choice is sampled by the underlying two choice models respectively. TCNet architecture and training process: Both trained TCNets contain a single-layer candidates encoder and an assortment encoder with one embedding sub-layer. The utility decoder also contains a single layer but without an activation function. We employ single-head attention, layer normalization, residual connection, and dropout (with dropout rate $0.1$) as described in \citep{vaswani2017attention}. The hidden dimensions of all sub-layers (if available) are set as $6$. The training process is the same as in Appendix \ref{apx:train_process} except there are only $20$ training epochs and we apply $0.002$ initial learning rate and $0.0005$ weight decay on the optimizer.

\end{document}